\documentclass{article}

\usepackage[preprint,nonatbib]{neurips_2020}

\usepackage[utf8]{inputenc} % allow utf-8 input
\usepackage[T1]{fontenc}    % use 8-bit T1 fonts
\usepackage{hyperref}       % hyperlinks
\usepackage{url}            % simple URL typesetting
\usepackage{booktabs}       % professional-quality tables
\usepackage{amsfonts}       % blackboard math symbols
\usepackage{nicefrac}       % compact symbols for 1/2, etc.
\usepackage{microtype}      % microtypography

\usepackage{graphicx}
\usepackage{amsmath}
\usepackage{amssymb}
\usepackage[shortlabels]{enumitem}
\usepackage[font={small}]{caption}
\usepackage[multiple]{footmisc}
\usepackage{multirow}
\usepackage{booktabs}
\usepackage[T1]{fontenc}
\usepackage[dvipsnames]{xcolor}
\usepackage{subfig}

\newcommand\jb[1]{\textcolor{blue}{[JB: #1]}}
\newcommand\ag[1]{\textcolor{violet}{[AG: #1]}}

\newcommand\bert{\textsc{BERT}}
\newcommand\genbert{\textsc{GenBERT}}

\newcommand\reformer{\textsc{Reformer}}
\newcommand\blockbert{\textsc{BlockBERT}}

\newcommand\comment[1]{}

\newcommand\squad{\textsc{SQuAD}}
\newcommand\hotpotqa{\textsc{HotpotQA}}

\newcommand\fone{$\mathrm{F}_1$}

\definecolor{myblue}{RGB}{60,120,216}
\definecolor{myred}{RGB}{221,126,107}

\usepackage{titlesec}
\titlespacing\section{0pt}{5pt plus 4pt minus 2pt}{3.33pt plus 2pt minus 2pt}
\titlespacing\subsection{0pt}{7pt plus 4pt minus 2pt}{4.66pt plus 2pt minus 2pt}

\usepackage{mathtools}
\usepackage{float}     % more table positioning options
\restylefloat{table}
\title{GMAT: Global Memory Augmentation for Transformers}

\author{%
  Ankit Gupta \\
  Tel Aviv University \\
  \texttt{ankitgupta.iitkanpur@gmail.com} \\
   \And
   Jonathan Berant \\
   Tel Aviv University, Allen Institute for AI \\
   \texttt{joberant@cs.tau.ac.il} \\
}

\begin{document}

\maketitle

% If you wish to post a preprint of your work online, e.g., on arXiv, using the
% NeurIPS style, please use the \verb+preprint+ option. This will create a
% nonanonymized version of your work with the text ``Preprint. Work in progress.''
% in the footer. This version may be distributed as you see fit. Please \textbf{do
%   not} use the \verb+final+ option, which should \textbf{only} be used for
% papers accepted to NeurIPS.

% At submission time, please omit the \verb+final+ and \verb+preprint+
% options. This will anonymize your submission and add line numbers to aid
% review. Please do \emph{not} refer to these line numbers in your paper as they
% will be removed during generation of camera-ready copies.

\setlength{\abovedisplayskip}{2pt}  % reducing whitespace
\setlength{\belowdisplayskip}{2pt}
\setlength{\belowcaptionskip}{-15pt}

\begin{abstract}
Transformer-based models have become ubiquitous in natural language processing thanks to their large capacity, innate parallelism and high performance. The contextualizing component of a Transformer block is the \emph{pairwise dot-product} attention that has a large $\Omega(L^2)$ memory requirement for length $L$ sequences, limiting its ability to process long documents. This has been the subject of substantial interest recently, where multiple approximations were proposed to reduce the quadratic memory requirement using sparse attention matrices.
In this work, we propose to augment sparse Transformer blocks with a dense attention-based \emph{global memory} of length $M$ ($\ll L$) which provides an aggregate global view of the entire input sequence to each position. Our augmentation has a manageable $O(M\cdot(L+M))$ memory overhead, and can be seamlessly integrated with prior sparse solutions. Moreover, global memory can also be used for sequence compression, by representing a long input sequence with the memory representations only. We empirically show that our method leads to substantial improvement on a range of tasks, including (a) synthetic tasks that require global reasoning, (b) masked language modeling, and (c) reading comprehension. 

%several approximations have been proposed for handling long inputs (e.g.\ Sparse Transformer, Reformer, etc). In this work, we propose to augment such sparse Transformer blocks with a dense attention-based \emph{global memory} of length $M$ ($\ll L$) which provides an aggregated global view of the entire input sequence to each position. Our augmentation 1) is orthogonal to the previous approximations and can be used in addition to them, 2) has a manageable $O(M\cdot(L+M))$ memory overhead, and, 3) leads to performance improvements on tasks such as language modeling, question answering, sequence tagging and sequence compression.
\end{abstract}

\section{Introduction}
\label{sec:intro}

The Transformer architecture \cite{vaswani2017attention} has been widely successful in achieving state-of-the-art performance on a wide range of natural language processing (NLP) tasks, including machine translation \cite{edunov2018understanding}, language modeling \cite{roy2020efficient}, question-answering \cite{karpukhin2020dense}, and many more. In particular, Transformers pre-trained on large amounts of text with a language modeling (LM) objective, have become the de-facto standard in NLP, exhibiting surprising amounts of linguistic and world knowledge \cite{peters2018elmo, devlin2018bert, liu2019roberta, lan2019albert, petroni2019language, hewitt2019structural,Roberts2020t5kb}. Moreover, Transformers with tailored attention patterns have been successfully used to replace convolutions in computer vision \cite{parmar18, ramachandran19}, and have also been useful in music generation \cite{huang2018music}, symbolic mathematics \cite{lample2019deep} and other modalities. 
% stablization of reinforcement learning \cite{RL stablization by Parisotto, et al. 2019}, etc. 

One of the most powerful features in Transformers is (pairwise) self-attention, where all positions in an input sequence aggregate information from the entire sequence in parallel. However, this requires computing a similarity score for all pairs of positions simultaneously, leading to a $\Omega(L^2)$ memory requirement for length $L$ sequences, which is prohibitively expensive for long sequences. To alleviate this issue, several \emph{sparsifications} of vanilla self-attention have been recently proposed; each restricting the number of positions that a given position can attend to \cite{yang2019xlnet,child2019generating,raecompressive2019,ye2019bp,qiu2019blockwise,kitaev2020reformer,beltagy2020longformer}. For example, in \blockbert{} \cite{qiu2019blockwise}, the sequence is split into $\frac{L}{M}$ chunks of length $M$ and positions in chunk $i$ only attend to positions in chunk $\sigma(i)$ for some pre-determined permutation $\sigma$, thereby having a $O(M\cdot L)$ memory requirement. The \reformer{} \cite{kitaev2020reformer} uses locality-sensitive hashing (LSH) to arrange similar vectors close to one another and then chunks them. Each chunk then attends to only a couple of chunks leading to a $O(M\cdot L)$ memory requirement. %\jb{Do you think it is better to have a separate `background' section where you explain about prior work on sparse stuff? or do you think it is enough here? It is ok, but I wonder if it is a bit too detailed for intro but still not clear enough?}
While such sparsifications often lead to performance that is comparable to vanilla Transformers, they have some undesirable consequences:
\begin{itemize}[leftmargin=*,topsep=0pt,itemsep=0pt,parsep=0pt]
    \item A position can require many layers to accumulate information from the entire input, and thus struggle when aggregating global information is necessary. For example, in \S\ref{sec:majority}, we show that a LSH Transformer (\reformer{} without reversible connections) struggles on a simple tagging task that requires information from the entire sequence.
    %position of a Boolean sequence by its mode. 
    \item Most sparsification schemes pose an inductive bias based on the locality of natural language, by restricting a position to only attend to its nearby tokens. While this is often a reasonable assumption, it raises several concerns. First, it is trivial to construct examples where locality is violated. For example, vanilla attention is invariant to input permutation, and thus can handle tasks such as ``word deshuffling'', where a randomly shuffled sequence needs to be mapped to the original order. On the other hand, any locality-based inductive bias would be detrimental. Second, progress in natural language understanding has led to increasing interest in handling global dependencies in long documents and even entire books \cite{kovcisky2018narrativeqa}, where a locality-based inductive bias is sub-optimal.
    %many tasks might violate it. For instance, consider the standard masked LM task used for training \bert{}-like models \jb{ref} where, given a partially masked sequence $I$, the model must map it to the original sequence $O$. On this task, the pre-training performance of a model with vanilla self-attention and learned positional embeddings would be \emph{no different} than on the task of mapping $\sigma(I)$ to $\sigma(O)$, for any fixed permutation $\sigma$. On the other hand, any locality-based inductive bias would be detrimental. Similarly, the ``word-salad'' task of mapping a jumbled input sequence to the original sequence does not satisfy this assumption. \jb{this argument is a bit weak right now, it seems you want to say this assumption is not always true, but then you give only made-up examples. You can say two things, first, that it is clear you can construct cases where sparsity is bad like you did. In addition, we want to go towards models that take longer and longer contexts into account}
    %\jb{The problem with the inductive bias:
%a) first of all it is clear that this inductive bias can lead to issues in some tasks, for example transformer with vanilla attention is fine with permutations and word salad but sparse transformers with locality bias will fail
%b) second, it is clear that from an applicative perspective in NLP we would like to go towards modeling global context more and more as is evident from recent interest in open-domain QA, QA over books, coref over multiple documents, etc.}
\end{itemize}

\begin{figure}[t]%\setlength{\belowcaptionskip}{-18pt}
    \centering
    \includegraphics[scale=0.54]{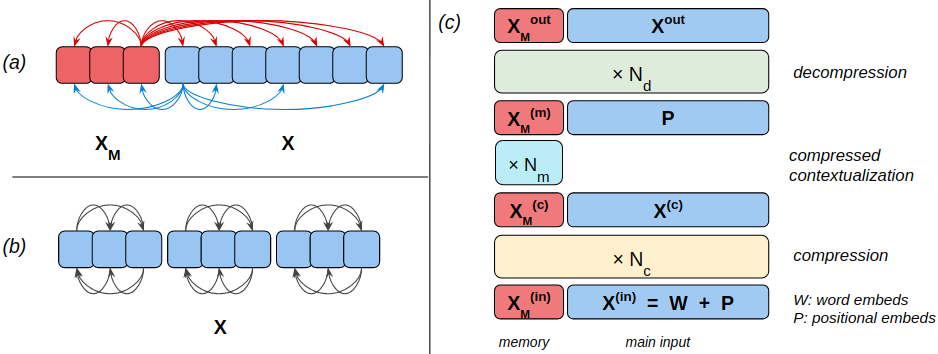}
    \caption{\textbf{(a)} \emph{GMAT}: For each attention head, tokens of the main sequence $X$ (blue) use \emph{approximate/sparse} attention for attending to $X$, and vanilla attention for attending to the global memory $X_M$. Tokens of $X_M$ (red) attend to all tokens using vanilla attention.  \textbf{(b)} \emph{chunked self-attention}: tokens of the main sequence use vanilla attention for attending to their respective chunk and global memory, but do not attend to other chunks.  \textbf{(c)} GMAT for sequence compression for a model with $N = N_c + N_m + N_d$ layers.}
    \label{figure:intro}
\end{figure}

In this work, we propose 
{\bf Global Memory Augmentation for Transformers (GMAT)}. 
%While it is desirable to have a model that works well without task-specific modifications, we believe that an important step in this direction would be to have an efficient Transformer variant that matches the perfovrmance of vanilla attention on various tasks.
%\jb{previous sentence is weird. how about ``In this work our goal is to develop an efficient transformer variant that matches the performance of vanilla attention on various tasks" - but - is this true? we do see degradation in performance}
We augment sparse variants of the Transformer with a small global memory which is read and updated by all the positions using vanilla attention.  
%Specifically (Figure~\ref{figure:intro}a), we prefix every input sequence $X$ (of length $L$) with a list of $M$ \emph{memory} tokens. At each multi-head attention layer, for each head, the $L$ tokens of the main sequence attend to other tokens of the main sequence using any sparse variant of attention, whereas they attend to the $M$ memory tokens using vanilla dense attention.
Specifically, we prefix every input sequence (of length $L$) with a list of $M$ \emph{memory} tokens. At each multi-head attention layer, for each head (Figure~\ref{figure:intro}a), the $L$ tokens\footnote{For brevity, we use the word \textit{token} to refer both to the input token and its contextualized representation interchangeably.} of the main sequence attend to other tokens of the main sequence using any sparse variant of attention, whereas they attend to the $M$ memory tokens using vanilla dense attention.
Moreover, the $M$ memory tokens attend to all $M+L$ tokens using vanilla attention. This results in a $O(M\cdot(L+M))$ memory overhead which is manageable for $M$ $\ll L$. Because the number of parameters in Transformers does not depend on the length of the input (modulo learned positional embeddings), the number of parameters grows by only a negligible $M\cdot E$ parameters, for an embedding size $E$.

We propose also to use GMAT for sequence compression (Figure~\ref{figure:intro}c). After encoding an input sequence with $N_c$ GMAT layers, we discard the vectors corresponding to the main sequence $X$, and keep only the global memory vectors, which are now a compressed representation of the entire input. The memory vectors are then processed and decompressed using $N_d$ layers back to the original input length. The sequence can now be stored using only $M (\ll L)$ vectors, and decompression is done with a small number $(N_d)$ of GMAT layers.

We evaluate GMAT on a wide range of tasks and show: (a) large improvements on synthetic tasks where global reasoning is required, (b) it improves masked langauge modeling (MLM) accuracy (used in Transformer pre-training), (c) improvements on two reading comprehension (RC) tasks, and last (d) moderate reductions in MLM and RC performance when using GMAT for compression.

To summarize, GMAT is a simple extension of the Transformers architecture, that can be seamlessly combined with any sparse attention variant. We show GMAT is useful for reducing memory requirements as well as for sequence compression, and demonstrate  performance enhancements on a wide range of tasks. Our code and data can be downloaded from \url{https://github.com/ag1988/gmat}.

\section{Global-Memory Augmented Transformers}\label{sec:model}

% \jb{
% formally explain it
% \begin{itemize}
%     \item usual transformer block
%     \item your transformer: it assumes chunks + memory + parameters for memory + positional embeddings
%     \item compression
%     \item paragraph on relation to longformer.
% \end{itemize}
% }

%Before formally describing our experimental set-up we recap some relevant definitions from \cite{vaswani2017attention}. 
A Transformer \cite{vaswani2017attention} is a stack of layers each consisting of sub-layers such as multi-head attention, feed-forward, etc. Its contextualizing component is the multi-head attention defined as follows.

\paragraph{Multi-head Attention} Given a query $Q \in \mathbb{R}^{L_Q \times d}$, key $K \in \mathbb{R}^{L_K\times d}$ and value $V \in \mathbb{R}^{L_K\times d}$, the output of scaled dot-product attention is defined as:
\begin{equation}\label{eqn:attention}
\mathrm{Attention}(Q,K,V) = \mathrm{softmax}\left(\frac{QK^T}{\sqrt{d}}\right)V .
\end{equation}
In multi-head attention, instead of computing a single attention output with $d_{\text{model}}$ dimensional keys, queries, and values, these are linearly projected down in parallel $h$ times to $d = d_{\text{model}} / h$ dimensions, using different learned projection matrices. Attention is applied to each of the $h$ new queries, keys and values, yielding $d$ dimensional outputs which are concatenated and again projected to obtain the $d_
\text{model}$-dimensional output.

The attention function (Eq. \ref{eqn:attention}) requires the computation of $QK^T$ containing $L_Q\cdot L_K$ entries and can be expensive for long sequences. To alleviate this issue, \textit{sparse attention} variants \cite{child2019generating, ye2019bp, qiu2019blockwise, kitaev2020reformer, beltagy2020longformer} relax this requirement and compute only a few entries of $QK^T$, masking out the rest. For a binary mask\footnote{The sparsity of $B$ can be leveraged via customized implementations of matrix product \cite{child2019generating,beltagy2020longformer}.} $B \in \{0,-\infty\}^{L_Q\times L_K}$,
\begin{equation}\label{eqn:sparse-attn}
\mathrm{SparseAttention}(Q,K,V,B) = \mathrm{softmax}\left(\frac{QK^T}{\sqrt{d}} + B\right)V .
\end{equation}

\paragraph{Global Memory} As explained in \S\ref{sec:intro}, sparse attention variants have some undesirable properties. To remedy this, we augment such models with a small global memory which is read and updated by all the positions using vanilla attention (Figure~\ref{figure:intro}a). Specifically, we prefix every token sequence $X$ (of length $L$) with a sequence of $M$ \emph{memory} tokens \texttt{[$m_1$]},..., \texttt{[$m_M$]}. At each multi-head attention layer of the model, at each head, the $L$ representations $X$ corresponding to the tokens of the main sequence attend to the other positions in $X$ using any sparse attention variant, but attend to the representations of \emph{all} memory tokens $X_M$ normally (Eq. \ref{eqn:gmat-attn}). Moreover, the memory tokens attend to all the $M+L$ tokens normally.
%\small
\begin{equation}\label{eqn:gmat-attn}
% \hspace{-2pt}
\widetilde{X} = \mathrm{SparseAttention}\left(X,\begin{bmatrix*}[l] X \\ X_M \end{bmatrix*},\begin{bmatrix*}[l] X \\ X_M \end{bmatrix*},\begin{bmatrix} B &  \\  & 0 \end{bmatrix}\right)
\ ,\ 
\widetilde{X}_M = \mathrm{Attention}\left(X_M,\begin{bmatrix*}[l] X \\ X_M \end{bmatrix*},\begin{bmatrix*}[l] X \\ X_M \end{bmatrix*} \right) .
\end{equation}
%\normalsize
%\jb{can we somehow write this with some formula using the attention notation to make sure there is no confusion? For the memory it is trivial to write that the memory after the attention is just computed with $M$ as query and $X$ as keys and values, but for the input $X$ it is a bit complex because of the sparse attention part? I think we should have something unambiguous here}
This results in a $O(M\cdot(L+M))$ memory overhead (manageable for $M$ $\ll L$). Moreover, this does not add \emph{any} parameters to the model, except for a negligible $M\cdot E$ parameters used to embed the $M$ new memory tokens with an embedding size of $E$.

\paragraph{Chunked self-attention} To explore the limits of GMAT and highlight its ability to contextualize over multiple fragments via a memory, we work with \emph{chunked self-attention} (Figure~\ref{figure:intro}b), a simple sparsification method. In $C\times k$ chunked self-attention (Figure~\ref{figure:intro}b), a sequence of length $C\cdot k$ is partitioned into $k$ contiguous chunks of length $C$. Each token within a given chunk uses vanilla (multi-head) attention to attend to tokens in its chunk in addition to the global memory but does not attend to other chunks. Hence, chunks interact with each other \textit{only} via the memory. Without memory, training with $C\times k$ attention is equivalent to training with vanilla Transformer over a length-$C$ sequence.
While more complex sparsification schemes are possible, this setup focuses on the ability to contextualize disjoint segments through the global memory only.
Note that a single model can be used with different values of $C$ and $k$, as Transformer models are invariant to the length of their input, aside for positional embeddings, which we handle below.
%JB: I think this sounds apologetic.
%We note that this set-up is not designed for obtaining state of the art results on LM but for investigating the abilities of a global memory.
We use the notation $(C\times k, M)$ to denote a chunked self-attention model where the input sequence $X$ has length $C\cdot k$ 
%and uses $C\times k$ chunked self-attention 
with a global memory of size $M$. 
%We will refer by \emph{input} and \emph{output} to the the main sequence inputs and output representations \emph{without} the global memory. \jb{do we need the last sentence? consider deleting}

\paragraph{Positional Embeddings} As attention is invariant to order, it is important to supply the model with positional information corresponding to the individual tokens. Rather than have a distinct learnable vector for each position \cite{vaswani2017attention}, we represent a position $p$ as a tuple $(q,r)$ where $r = p\ (\text{mod } 512)$, and $q = {\lfloor p / 512 \rfloor}$. Each $0 \leq r < 512$ and $0 \leq q < 64$ has a distinct learnable vector.\footnote{These particular values allow us to initialize the vectors for $r$ with the learned 512 positional embeddings of pre-trained LMs such as \bert{}.} The positional embedding of $p$ is represented by the sum of the vectors corresponding to $q$ and $r$ and allows us to model positions up to $2^{15}$. Memory tokens have a fixed position, and thus positional embeddings are used only for the main sequence $X$.

\subsection{Sequence Compression}\label{sec:seq-compression}

Contextualized word representations  \cite{peters2018elmo,devlin2018bert} improve performance compared to fixed word embeddings such as GloVe \cite{pennington2014glove}.
%JB: for simplicity
%even without backpropogating the gradients through the contextualizing layers.
Unfortunately, some large models that compute contextualized representations do not fit on popular GPUs and need specialized hardware \cite{raffel2019exploring}.
Instead of leaving this computation to the users, an appealing option might be to release pre-computed contextualized representations, at least for popular benchmarks, similar to word embeddings \cite{pennington2014glove}. However, storing a vector for each position in a large corpus is expensive. A second natural motivation for GMAT is for \emph{sequence compression}, i.e. using a small memory to represent a long sequence. This can dramatically reduce the overhead in storing pre-computing contextualized representations for large corpora.

% Contextualized word representations \cite{peters2018elmo,devlin2018bert} are known to outperform fixed word embeddings such as GloVe, etc \cite{pennington2014glove} even without backpropogating the gradients through the contextualizing layers. Unfortunately, while larger LMs lead to more informative respresentations, some of these models are too large to fit on popular GPUs and need specialized hardware even for the forward pass \cite{raffel2019exploring}.
%Instead of leaving such a computation to the users, one option might be to release pre/partially computed representations, at least for the popular benchmarks, similar to \cite{pennington2014glove}. As the basic layer operations in a Transformer (self-attention, feed-forward, etc) are length-preserving, storing a vector for each position can be expensive. While thus far, the motivation for GMAT has been to reduce run-time memory requirements, it can be naturally used for \emph{sequence compression} i.e. using the small global memory to represent an entire long sequence, and serialize it in a compressed form for later use.

\comment{Thus far, the motivation for GMAT has been to reduce run-time memory requirements. However, GMAT can also be naturally used for \emph{sequence compression}, that is, using the small global memory to represent an entire long sequence, and serialize it in a compressed form for later use.}

Consider a $N$-layer GMAT with a memory of size $M$. We apply the $N$ model layers in 3 steps as shown in Figure~\ref{figure:intro}c. Given a length $L$ input sequence, let $W$ and $P$ denote its word and positional embeddings. First, the bottom $N_c$ layers are applied for \textit{compressing} the entire information of the input sequence into just $M$ memory vectors $X_M^{(c)}$. The next $N_m$ layers are then applied only on the $M$ memory vectors resulting in richer representations $X_M^{(m)}$.
%JB: just to make shorter, don't think it is very important.
%and saving run-time computation. 
This length $M$ sequence is restored to the original length $M+L$ by concatenating the positional embeddings $P$ and, finally, the remaining $N_d=N-N_c-N_m$ layers are applied for \textit{decompressing} the information packed in $X_M^{(m)}$ into the final representations. Here, the positional embeddings $P$ act as queries for restoring the original input information from $X_M^{(m)}$.

The $M (\ll L)$ vectors $X_M^{(m)}$ can be efficiently serialized for later use and are decompressed using minimal post-processing into representations of length $L$. In \S\ref{sec:compress_eval} and \S\ref{sec:rc}, we show that using the decompressed representations leads to only a small performance degradation on masked language modeling and reading comprehension.
%that perform on a par with vanilla-contextualized representations (\S\ref{sec:compression}). 
We use positional embeddings $P$ in the decompression step and \emph{not} the contextualized representation $X^{(c)}$ (Figure~\ref{figure:intro}c), since we want the output to depend only on $M$ vectors instead of $M+L$.

\paragraph{Comparison to Longformer}
In contemporaneous work \cite{beltagy2020longformer}, the authors demonstrate the effectiveness of a sparse sliding window attention pattern combined with a global memory on multiple natural understanding tasks. In contrast to our work, the contribution of the global memory component is not evaluated, and is designed on a task-by-task basis.
Moreover, attention scores for the memory and the main sequence are computed using separate sets of projection matrices, thereby introducing new parameters. In comparison, we demonstrate the utility of a global memory for various sparse attention patterns on synthetic, NLP and compression tasks, introducing a minimal set of new parameters.

\comment{
\paragraph{Memory Augmented Models} The past work on memory augmented neural networks is mostly focused towards mitigating the limitations of recurrent neural networks \cite{graves2014neural,weston2015memory,sukhbaatar2015end,grave2016improving}. In the context of Transformers, in contemporaneous work \cite{beltagy2020longformer}, the authors demonstrate the effectiveness of a sparse sliding window attention pattern combined with a global memory on multiple natural understanding tasks. In contrast to our work, the contribution of the global memory component is not evaluated, and is designed on a task-by-task basis.
Moreover, attention scores for the memory and the main sequence are computed using separate sets of projection matrices, thereby introducing new parameters. In comparison, we demonstrate the utility of a global memory for various sparse attention patterns on synthetic, NLP and compression tasks, introducing a minimal set of new parameters.
}

\section{Global Reasoning on Synthetic Data}\label{sec:synthetic}

%JB: delete for brevity?
We consider two synthetic datasets, where global reasoning over the entire input is required.

\subsection{Majority Tagging}\label{sec:majority}
Recently proposed sparse attention schemes exploit the locality of natural language (\S\ref{sec:intro}). One exception is locality sensitive hashing (LSH) attention \cite{kitaev2020reformer}. While in LSH attention, tokens are not bound to attend only within their proximity, it does limit the number of positions a token can attend to at each head. We examine the utility of adding GMAT to a LSH Transformer \cite{kitaev2020reformer}, that is, a Transformer that uses LSH attention, for a sequence tagging task.

\textit{$\mathrm{MAJORITY}(L, p)$}: Let $X = (x_1,\ldots,x_L)$ be a sequence of $L$ integers from the set $\{1,\ldots,2p\}$. Let $N_i$ denote the number of occurrences of $i$ in $X$. For every \textit{even} integer $i$ in the domain, define
$$ \mathrm{maj}(i-1) = \mathrm{maj}(i) = \begin{cases}
i - 1  & N_{i - 1} \ge N_{i} \\
i & \text{otherwise}
\end{cases}. $$
Then the \textit{majority sequence} is defined to be $(\mathrm{maj}(x_1),\ldots,\mathrm{maj}(x_L))$. Note that for $p = 1$, the task requires tagging all tokens of $X$ by their mode.

\begin{figure}[t]
    \centering
    \subfloat[2-layers]{\label{figure:majority}\includegraphics[scale=0.35]{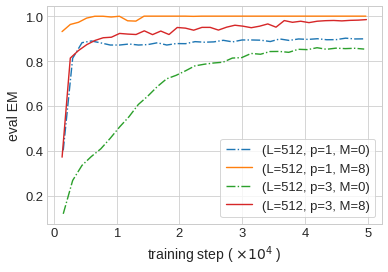}}
    \hspace*{0.5in}\subfloat[12-layers]{\label{figure:majority12}\includegraphics[scale=0.35]{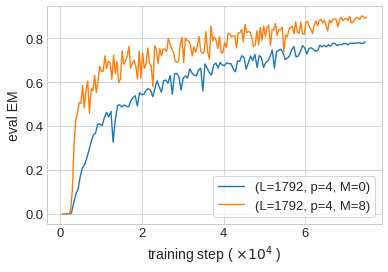}}
    \caption{Evaluation exact match (EM) of LSH Transformer on the $\mathrm{MAJORITY}(L, p)$ task of \S\ref{sec:majority}. $M$ denotes the memory size. Models in the same figure use the same hyperparameters.}
\label{figure:maj}
\end{figure}

To create the data, we sampled the elements of $X$ independently and uniformly from the domain. After training a 2-layer LSH Transformer on the above task (Figure~\ref{figure:majority}) we compared its performance with and without a global memory of size $M = 8$. Hyperparamaters for LSH attention (bucket size, rounds of hashing, etc.) were used as suggested by \cite{kitaev2020reformer} and the other hyperparameters (hidden size, etc.) were taken from  \bert{}-base (without dropout). As shown in Table~\ref{table:majority}, we found that LSH attention failed to do well on the $\mathrm{MAJORITY}(8192, 1)$ task. On much shorter inputs of length 512, it managed to do well on $\mathrm{MAJORITY}(512, 1)$ but again struggled on a slightly more complex task of $\mathrm{MAJORITY}(512, 3)$. Conversely, GMAT obtains near perfect performance on these tasks.

\begin{table}[h!]%\setlength{\belowcaptionskip}{-5pt}
    \scriptsize
    \centering
\begin{tabular}{|c|c|c|c|}\hline
  & $\mathrm{MAJORITY}(8192, 1)$ & $\mathrm{MAJORITY}(512, 1)$ & $\mathrm{MAJORITY}(512, 3)$ \\\hline
$M = 0$        & 0.15  & 0.92  & 0.86    \\\hline
$M = 8$ & \textbf{0.98}  & \textbf{1.0}  & \textbf{0.98} \\\hline
\end{tabular}
    \caption{Evaluation exact match (EM) of a 2-layer LSH Transformer on majority tagging. EM for an example is 1 iff every position is tagged correctly. $M$ denotes the memory size. %Training in the same column use the same hyperparamters except $M$.
    }
    \label{table:majority}
\end{table}

To determine if GMAT also leads to lower sample complexity in deeper models, we trained (Figure~\ref{figure:majority12}) a 12-layer model on $\mathrm{MAJORITY}(1792, 4)$ and found GMAT obtains better performance with lower sample complexity. This suggests that in LSH attention, while a certain level of sparsity  works well when information is mostly local, it can be too restrictive for inherently global tasks.

% \begin{figure}%\setlength{\belowcaptionskip}{-18pt}
%     \centering
%     \includegraphics[scale=0.4]{figures/majority.png}
%     \caption{2-layer LSH Transformer on $\mathrm{MAJORITY}(L, p)$ task of \S~\ref{sec:synthetic}. Trainings use the same hyperparameters except the ones shown.}
%     \label{figure:majority}
% \end{figure}

%-----------------------------------------------------------------------
\subsection{Numerical Reasoning Over Text}\label{sec:syntext} 

Having shown the utility of GMAT on a combinatorial task, we now transition towards language tasks. 
%As we will see later in \S\ref{sec:hotpot}, NLP datasets can have reasoning shortcuts allowing the models to answer questions without actually aggregating information from the entire context \cite{min2019compositional,jiang2019avoiding}.
We create a pseudo-textual task that requires global mathematical reasoning by generating examples from a rule-based generator from \cite{ggb2020injecting}.
%\ag{annonymize? \jb{but it is public, I think no one will check if the generator itself is public and we should it make it public anyway soon}}.
The generator generates examples that include a passage $P$, describing a sequence of related events, and a question $Q$ that requires aggregating information from various parts of $P$ and performing numerical reasoning to arrive at a numeric answer $A$ (see Table~\ref{table:syntext_example}).

%To ensure that our task indeed requires multi-hop reasoning, we synthesize pseudo-textual passage-question-answer triples using the set-up of \cite{ggb2020injecting}. As shown in Table~\ref{table:syntext_example}, each sample consists of a passage $P$ describing a sequence of related events and a question $Q$ that requires aggregating information from various parts of $P$ and performing numerical reasoning to arrive at a numeric answer $A$. 

\begin{table}[h!]
\begin{center}
\footnotesize
\begin{tabular}{p{13.5cm}}
\toprule
\textbf{P}: \textit{The \textbf{\textcolor{myblue}{commander}} recruited 358 households and \textbf{\textcolor{myred}{9669}} \textbf{\textcolor{myblue}{Italian troops}}. The commander lost 812 of the households. The commander recruited 542 households in France. The commander recruited 3075 households in France . The commander recruited 2843 households and \textbf{\textcolor{myred}{5413}} \textbf{\textcolor{myblue}{Native American troops}}.} 
\\ \hline
\textbf{Q}: How many \textbf{\textcolor{myred}{more}} \textbf{\textcolor{myblue}{Italian troops}} did the \textbf{\textcolor{myblue}{commander}} have than \textbf{\textcolor{myblue}{Native American troops}}? \qquad \textbf{A}: \textbf{\textcolor{myred}{4256}}
\\% \\ \hline
\bottomrule
\end{tabular}
\end{center}
\caption{An example from the textual synthetic data used in \S\ref{sec:syntext}.}
\label{table:syntext_example}
\end{table}

Following \cite{ggb2020injecting}, we train a generative encoder-decoder model, \genbert{}, on the generated data after replacing the encoder self-attention \cite{vaswani2017attention} with chunked self-attention (\S\ref{sec:model}), and compare the performance before and after GMAT augmentation (see training and data details in \S\ref{sec:syntext_data}). We summarize the results in Table~\ref{table:syntext}. Compared to vanilla attention over the entire input (i.e., $(140\times 1, 0)$), chunking the encoder input into 2 chunks significantly reduced the performance (i.e., $(70 \times 2, 0)$), in line with the global nature of the task. Surprisingly, adding a global memory of size 30 reduced accuracy even further. We hypothesize this is due to the strict weight-tying strategy employed by \genbert{}, where the parameters of the Transformer encoder and decoder are tied, leading to underfitting. To handle that, we untie the parameters of the projection matrices that update the memory representations $X_M$ in all attention heads (Eq.~\ref{eqn:gmat-attn}, right), initializing them randomly. \comment{Q, K, V are not parameters but outputs of projection} This separates the attention heads that update the main sequence from the heads that update the memory. In this setup, accuracy improved substantially, almost recovering the performance of vanilla attention.

\begin{table}[h!]%\setlength{\belowcaptionskip}{-5pt}
    \scriptsize
    \centering
\begin{tabular}{|c|c|c|c|}\hline
$(70\times 2, 0)$ & $(70\times 2, 30)$ & $(70\times 2, 30)$ (untied) & $(140\times 1, 0)$ \\\hline
62.2 & 51.4 &  86.7 & \textbf{93.9}    \\\hline
\end{tabular}
    \caption{Evaluation exact match (EM) of \genbert{} on the textual synthetic data. EM for a sample is 1 iff every digit of the desired number is predicted correctly.
    All models use same hyperparamters.}
    \label{table:syntext}
\end{table}

\section{Masked Language Modeling}\label{sec:lm}

One of the biggest success stories of Transformers is as an architecture for pre-training LMs. We now investigate pre-training GMAT with a masked language modeling objective, as a memory-efficient replacement for models such as \bert{} \cite{devlin2018bert, liu2019roberta}. Past work has shown strong correlation between performance on the MLM task and that on downstream applications \cite{liu2019roberta, lan2019albert}. For our experiments, we use the \emph{\bert{}-base} architecture \cite{devlin2018bert} after making the modifications described in \S\ref{sec:model}. 

We form examples by sampling sequences of length $L$ from English Wikipedia and the PG19 dataset, and replacing sub-words with the \texttt{[MASK]} token following the procedure in \cite{devlin2018bert} (details in \S\ref{sec:mlm_data}). The model is trained to maximize the log probability of the masked out tokens. We evaluate the \emph{error} of the model as the fraction of tokens predicted incorrectly, and the \emph{MLM ``perplexity''} as the reciprocal of the geometric mean of probabilities of all masked out tokens.\footnote{Equivalently, the natural exponential of the average loss over the development set.} %$\left(\prod_{t\in \mathcal{M}}\Pr[t]\right)^{-1/|\mathcal{M}|}$ 
PG19 contains 29K long books, and is thus likely to benefit from modeling long context, while in Wikipedia most articles are short and can fit into the 512 word pieces that are the input of \bert{}. We experiment with training a Transformer from scratch, as well as initializing with \bert{}-base.

\subsection{Random Initialization}\label{sec:rand_init}
\begin{figure}[t]
    \centering
    \hspace*{-0.1in}\subfloat[Wikipedia (random init)]{\label{figure:rand_init_wiki}\includegraphics[scale=0.35]{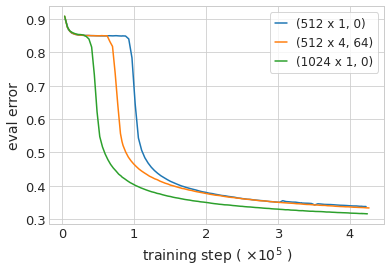}}
    \hspace*{-0.05in}\subfloat[Wikipedia (\bert{} init)]{\label{figure:bert_init_wiki}\includegraphics[scale=0.35]{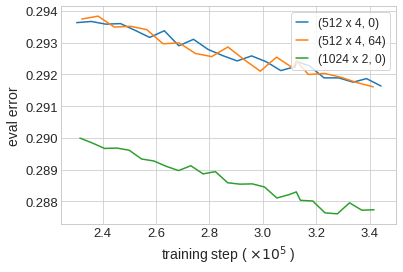}}
    \hspace*{-0.05in}\subfloat[PG19 (\bert{} init)]{\label{figure:bert_init_pg19}\includegraphics[scale=0.35]{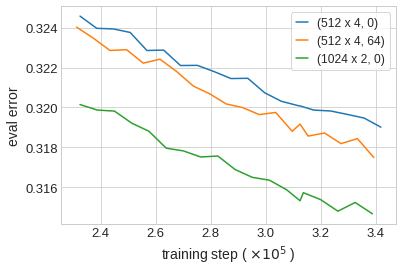}}
    \caption{Evaluation error for the MLM task in 3 different setups.}
\label{figure:lm}
\end{figure}

As shown in Figure~\ref{figure:rand_init_wiki}, we train 3 models on Wikipedia. The setting $(512\times 1, 0)$ corresponds to standard MLM training on instances of length 512 without global memory. Similarly, $(1024\times 1, 0)$ denotes training with vanilla attention over a context of size 1024, incurring a large memory penalty.\footnote{We do not train in the  $(2048\times 1, 0)$ setting due to memory constraints.} Lastly, in $(512\times 4, 64)$, a 2048-long context is chunked into four 512-chunks that have to interact via a global memory of size 64. Increasing the context size to 1024 improves MLM performance on Wikipedia (Table~\ref{table:rand_init}). Using global memory improves sample complexity and performance compared to training on 512-long instances, albeit only moderately. Thus, the chunks are able to leverage global memory to exchange information,  alleviating the context fragmentation problem \cite{yang2019xlnet}.

%We observed that contextualizing over a larger context of size 1024 significantly improves the MLM performance on the Wikipedia instances. Moreover, a global memory leads to lower sample complexity compared to the conventional training on 512-long instances (Table~\ref{table:rand_init}). The chunks are able to leverage the global memory for exchanging information among themselves thereby elevating the context fragmentation problem \cite{yang2019xlnet} to a certain extent. \jb{this is not very impressive, we should think if to keep or delete}zx

\begin{table}[h!]\setlength{\tabcolsep}{8pt} 
    \scriptsize
    \centering
    \begin{tabular}{|c|c|c|c|}\hline
    setting  & $(512\times 1, 0)$ & $(512\times 4, 64)$ & $(1024\times 1, 0)$ \\\hline
    best evaluation error / perplexity & 33.67 / 5.14 & 33.25 / 5.11 & \textbf{31.53 / 4.64}  \\\hline
    \end{tabular}
    \caption{MLM training on Wikipedia from random initialization for 431K steps. Error denotes the percentage of masked tokens predicted incorrectly.}
    \label{table:rand_init}
\end{table}

%We formed examples for the MLM using English Wikipedia and the PG19 datasets \cite{raecompressive2019} (details in\S\ref{sec:mlm_data}). \jb{I think some details from appendix should be moved here} While Wikipedia has several short articles, PG19 is comprised of books each containing thousands of tokens \jb{it is not clear why you are writing this sentence}. To investigate the utility of a global memory, we conduct separate MLM experiments using both random initialization and \bert{} initialization and ensure that the compared trainings are indeed comparable in terms of the hyperparameters such as learning rate, number of tokens in a batch, etc. \jb{this paragraph is confusing, just a bunch of unrelated sentences strung together}

\subsection{\bert{} Initialization}\label{sec:bert_init}
To show that GMAT can be easily integrated into existing pre-trained LMs, we take a pre-trained \bert{}-base model, and further train it using GMAT. Because \bert{} was pre-trained on Wikipedia, improving performance on Wikipedia itself could be difficult, as it already has high confidence on tokens from this corpus. Hence, we also train on PG19, which was not part of \bert{}'s training data.

%jb: i found this paragraph problematic.
%Although training from random initialization demonstrates the utility of GMAT on the MLM task, it is not immediately clear if the improvement will still be maintained if we start training from an already pre-trained LM such as \bert{}, especially given that it was already trained on Wikipedia. If the model can already predict the masked tokens well using only a context of size 512 there will not be enough training signal left for the global memory. To partly elevate this issue, we also experiment with the PG19 dataset which was not a part of \bert{}'s training data.
%Secondly, starting from pre-trained LMs also help us highlight the versatility of GMAT and that it can be easily integrated with the pre-trained LMs containing a plethora of linguistic information. 

Table~\ref{table:bert_init} summarizes the results. On Wikipedia, increasing the context size to 1024 provides a significant improvement (Figure~\ref{figure:bert_init_wiki}), but global memory does not improve performance compared to standard MLM training on 512-long instances. However, on PG19 (Figure~\ref{figure:bert_init_pg19}) using global memory substantially improves perplexity from $4.4 \rightarrow 4.35$, closing roughly half the gap from using a context of size $1024$, which obtains an MLM perplexity of $4.3$.  
This hints that the lack of  improvement on the Wikipedia data might be due to the fact that \bert{} was pre-trained on Wikipedia. 

\begin{table}[h!]%\setlength{\belowcaptionskip}{-5pt}
\setlength{\tabcolsep}{4pt} %% default is 6pt
    \scriptsize
    \centering
\begin{tabular}{|c|c|c|c|c|c|c|c|c|}\hline
 setting & \bert{} (no training) & $(512\times 4, 0)$ & $(512\times 4, 64)$ & $(1024\times 2, 0)$ & $(8\times 64, 0)$ & $(8\times 64, 64)$  \\\hline
 %initialization & & & & & $(512\times 4, 0)$ & $(512\times 4, 64)$ \\\hline
Wikipedia & 35.2 / 6.856  & 29.163 / 3.953   & 29.15 / 3.956  & \textbf{28.74 / 3.87} & 53.11 / 20.68 & 32.98 / 4.94  \\\hline
PG19 & 42.5 / 10.57 & 31.90 / 4.40 & 31.72 / 4.35 & \textbf{31.44 / 4.30} & -  & - \\\hline
\end{tabular}
    \caption{Evaluation error / perplexity for MLM training.. Models are initialized with \bert{}-base, except for  $(8\times 64, 0)$, $(8\times 64, 64)$, which are initialized with the trained models $(512\times 4, 0)$, $(512\times 4, 64)$ respectively.}
    \label{table:bert_init}
\end{table}

The above results indicate that disjoint text segments can exchange useful information via global memory.
However, because natural language has a locality bias, the utility of memory diminishes as the chunk length $C$ increases.
To determine the efficacy of GMAT when $C$ is small, where contextualization should highly depend on the memory, we experiment with chunks of size 8. As expected, without access to a reasonably-large surrounding context, the model $(8\times 64, 0)$ fails to predict masked tokens (Table~\ref{table:bert_init}). Interestingly, a small global memory of size $64$ significantly improves performance ($53.11 \rightarrow 32.98$ error, $20.68 \rightarrow 4.94$ perplexity), reaching performance that is close to $(512\times 4, 0)$. We further evaluate the pre-trained GMAT models on reading comprehension tasks in \S\ref{sec:rc}.

\subsection{Sequence Compression}\label{sec:compress_eval}

We turn to sequence compression, where our goal is to compress a sequence of length $L$ into $M$ vectors that can be saved and later decompressed back into a sequence of length $L$, with minimal drop in performance. Using the setup described in \S\ref{sec:seq-compression}, we use $N_c$ compression layers, followed by $N_d = N - N_c$ decompression layers, and train with the same data and MLM objective as above on Wikipedia.
As shown in Table~\ref{table:compression}, we found that $N_c=9$ outperforms $N_c=3$ (which also happens to be well-aligned with our need for a small number of decompression layers). Compared to a model without compression, we observe a moderate degradation in performance
%\footnote{we conjecture this can be further improved via a more carefully chosen dropout profile.}
($29.163 \rightarrow 32.98$ error, and $3.953 \rightarrow 5.017$ MLM perplexity), showing that a global memory of size just 64 provides a compact and useful representation for the entire sequence of length 512.

\begin{table}[h!]%\setlength{\belowcaptionskip}{-5pt}
    \scriptsize
    \centering
\begin{tabular}{|c|c|c|c|c|c|c|c|c|}\hline
  setting & $(512\times 4, 0)$ & \begin{tabular}[c]{@{}c@{}} $(512\times 1, 64)$,  $N_c=3$
 \end{tabular} & \begin{tabular}[c]{@{}c@{}} $(512\times 1, 64)$, $N_c=9$\end{tabular} \\\hline
initialization & \bert{} & $(512\times 4, 64)$ & $(512\times 4, 64)$ \\\hline
best evaluation error / perplexity & 29.163 / 3.953  & 33.44 / 5.112 & 32.98 / 5.017 \\\hline
\end{tabular}
    \caption{MLM training on Wikipedia with compression. Compressed models were initialized with the $(512\times 4, 64)$ model trained in \S\ref{sec:bert_init} and further trained for 440K steps.}
    \label{table:compression}
\end{table}

\subsection{Reading Comprehension Performance}\label{sec:rc}

While MLM performance is known to correlate well with downstream applications \cite{liu2019roberta, raffel2019exploring}, we take Wikipedia-based GMAT models trained with the MLM objective in \S\ref{sec:bert_init} and \S\ref{sec:compress_eval} and further fine-tune them on reading comprehension (RC) tasks. 

\paragraph{\squad{}} We first fine-tune on \squad{} v1 \cite{rajpurkar2016squad}, using the simple sliding-window based approach of \cite{devlin2018bert}. Similar to past work \cite{lan2019albert}, we limit the input size to 384 tokens, as most paragraphs are relatively short. We train all models using identical hyperparameters. Summarized in Table~\ref{table:squad}, the model $(512\times 4, 64)$ improves performance over \bert{} ($88.6 \rightarrow 89.2$ \fone{}), indicating global memory helps even with vanilla self-attention. The performance of $(512\times 4, 0)$ is very similar to \bert{}, ruling out the possibility that the performance of $(512\times 4, 64)$ was a result of extra pre-training on Wikipedia.
Surprisingly, the model $(8\times 64, 64)$ reported $84.2$ \fone{}, a moderate drop in performance given that, with chunks of size 8, the contextualization depends almost entirely on the memory. Interestingly, the compression model with $N_c=9$ reported $87.1$ \fone{} (compared to \bert{}'s $88.6$) an impressive score given that, after 9 layers, the information of the entire input must pass through only 64 vectors.

\begin{table}[h!]\setlength{\tabcolsep}{5.3pt} %% default is 6pt
    \scriptsize
    \centering
\begin{tabular}{|c|c|c|c|c|c|c|}\hline
   \bert{} & $(512\times 4, 0)$ & $(512\times 4, 64)$ & $(8\times 64, 64)$ & $(8\times 64, 0)$ & \begin{tabular}[c]{@{}c@{}} $(512\times 1, 64)$\\ $N_c=3$\end{tabular} &  \begin{tabular}[c]{@{}c@{}} $(512\times 1, 64)$\\ $N_c=9$\end{tabular} \\\hline
 81.09 / 88.60 & 80.93 / 88.47   & \textbf{81.77 / 89.16}  & 75.64 / 84.17  & 9.82 / 14.60  & 76.59 / 84.58               & 79.55 / 87.05 \\\hline
\end{tabular}
    \caption{Evaluation EM/\fone{} on \squad{} v1.
    %while finetuning the Wikipedia models from \S~\ref{sec:bert_init}, \S~\ref{sec:compression}. All finetunings use the same relevant hyperparameters.
    }
    \label{table:squad}
\end{table}

\paragraph{\hotpotqa{}} To investigate the utility of GMAT for long-range reasoning, we fine-tune our models on \hotpotqa{} \cite{yang2018HotpotQAAD}, an RC dataset focusing on multi-paragraph reasoning. In \hotpotqa{}, examples comprise of a question $Q$, 2 gold paragraphs $G_1, G_2$ required for answering $Q$, and 8 distractor paragraphs. Each gold paragraph contains \emph{supporting facts}: the sentences relevant for answering $Q$.

As we are interested in analyzing if models can aggregate information dispersed in a long context, we order the paragraphs such that one of $G_1, G_2$ is among the first 3 paragraphs, and the other is among the last 3. We refer to this setting as the ``gold-seperated'' setting,  denoted by subscript \textsc{GS}. To reuse the sliding-window setup from \squad{}, we concatenate the 10 paragraphs (details in \S\ref{sec:hotpot-details}) into a single long context $D$. Following \cite{devlin2018bert}, each input instance consists of $Q$ concatenated with a window $P$ of contiguous text from $D$ whose length is limited according to the maximum sequence length allowed by the model (512 for \bert{}, 2048 for $(512\times 4, 0)$ and $(512\times 4, 64)$). We leverage the supporting facts supervision provided in the dataset, and include a binary tagging task (denoted by \textsc{SF}) of independently tagging each token of the input by 1/0 indicating if it is part of a supporting fact (\S\ref{sec:hotpot-details}). Moreover, for GMAT models we replaced the input memory tokens by the first 64 tokens of $Q$, as this improved performance. 

\begin{table}[h!]\setlength{\tabcolsep}{14pt}
    \scriptsize
    \centering
\begin{tabular}{|c|c|c|c|c|}\hline
model  & \multicolumn{2}{c|}{\bert{}} & \multicolumn{2}{c|}{$(512\times 4, 64)$} \\\hline
\textsc{SF} task included & no & yes & no & yes \\\hline
all & 66.3 & 66.3 & 67.6 & \textbf{68.3} \\\hline
only-comparison & 61.6 & 58.2 & 65.3 & \textbf{65.6} \\\hline
\end{tabular}
    \caption{\fone{} scores on $\hotpotqa{}_{\textsc{GS}}$ development set.} \label{table:hotpot}
\end{table}

The results after finetuning on $\hotpotqa{}_{\textsc{GS}}$ are summarized in Table~\ref{table:hotpot}. With the ability to contextualize over a much larger context containing both $G_1,G_2$, the model $(512\times 4, 64)$ reported an improved performance compared to \bert{} ($66.3 \rightarrow 68.3$ \fone{}). The improvement on the ``comparison'' type of questions is even more pronounced ($61.6 \rightarrow 65.6$) as such questions usually require comparing quantities from each $G_1, G_2$ and are hard if either of $G_1, G_2$ is missing from the input. We omit results for $(512\times 4, 0)$, because chunked self-attention without memory means that the chunk that contains the question has no access to the entire context, leading to low performance.
%as, due to chunked self-attention, the question (present in the first 512-chunk) does not have access to the entire context leading to significant performance degradation.

To fairly compare $(512\times 4, 0)$ and $(512\times 4, 64)$ and study the effect of global memory, we create a new setting, ``question-repeated'' (\textsc{QR}), in which, while forming the context $D$, $Q$ is appended after each paragraph thereby ensuring that all four 512-chunks of $(512\times 4, 0)$ have access to $Q$. Moreover, we repeat this experiment after creating an adversarial version \textsc{ADV-}\hotpotqa{} of \hotpotqa{} by following \cite{min2019compositional}. The results after fine-tuning on these datasets are summarized in Table~\ref{table:hotpot-qr}. Global memory improves model performance on both $\hotpotqa{}_{\textsc{GS}+\textsc{QR}}$ ($66.4 \rightarrow 67.4$ \fone{}) and $\textsc{ADV-}\hotpotqa{}_{\textsc{GS}+\textsc{QR}}$ ($64 \rightarrow 64.6$ \fone{}) reaffirming its utility for long-range reasoning.

\begin{table}[h!]\setlength{\tabcolsep}{13pt}
    \scriptsize
    \centering
\begin{tabular}{|c|c|c|c|c|}\hline
model  & \multicolumn{2}{c|}{$(512\times 4, 0)$} & \multicolumn{2}{c|}{$(512\times 4, 64)$} \\\hline
\textsc{SF} task included & no & yes & no & yes \\\hline
$\hotpotqa{}_{\textsc{GS}+\textsc{QR}}$ & 65.4 & 66.4 & 66.0 & \textbf{67.4} \\\hline
$\textsc{ADV-}\hotpotqa{}_{\textsc{GS}+\textsc{QR}}$ & 62.6 & 64.0 & 64.0 & \textbf{64.6} \\\hline
%only-comparison & 64.0 & 58.2 & 65.3 & \textbf{65.6} \\\hline
\end{tabular}
    \caption{\fone{} on $\hotpotqa{}_{\textsc{GS}+\textsc{QR}}$ and $\textsc{ADV-}\hotpotqa{}_{\textsc{GS}+\textsc{QR}}$ development sets.}
    \label{table:hotpot-qr}
\end{table}

\section{Conclusion}

In this work, we proposed GMAT, a simple extension to the Transformer architecture that allows a better trade-off between compute and performance and can be naturally used for sequence compression. Our approach can be seamlessly integrated with the increasingly-popular sparse Transformer variants. We show GMAT (a) leads to performance improvements on a wide range of tasks, and (b) can be used to compress long sequences by factor of $8\times$ with only a small degradation in performance.

\newpage
\section*{Broader Impact}
Transformers have become a popular architecture for sequence processing and generation in natural language processing and outside of it. The goal of this paper it to reduce the memory requirement and thereby allow for longer sequences to be processed. Moreover, our compression technique can facilitate the use of pre-computed contextualized representations, allowing users access to an approximation of these representations even if they cannot compute the representations from scratch themselves. As such, we consider a positive impact of this work to be the ability of more users with constraints on their computational resources to use the Transformer architecture and its pre-trained representations. Moreover, being able to process long documents can open the door to new applications in natural language processing, such as multiple-document understanding, and perhaps also processing of sequences outside of NLP, for example in Biology. As Transformers are becoming ubiquitous in machine learning, naturally any negative impact that can be attributed to Transformers (fake news generation, classifiers in sensitive domains such as the justice system and healthcare) are also inherited by our approach, and perhaps enhanced when long sequences need to be processed.

\begin{ack}
We thank Shimi Salant, Yoav Goldberg and Mor Geva for helpful discussions and constructive suggestions. This research was partially supported by The Israel Science Foundation grant 942/16, The Yandex Initiative for Machine Learning, and the European Research Council (ERC) under the European Union Horizons 2020 research and innovation programme (grant ERC DELPHI 802800).

% Use unnumbered first level headings for the acknowledgments. All acknowledgments
% go at the end of the paper before the list of references. Moreover, you are required to declare 
% funding (financial activities supporting the submitted work) and competing interests (related financial activities outside the submitted work). 
% More information about this disclosure can be found at: \url{https://neurips.cc/Conferences/2020/PaperInformation/FundingDisclosure}.

% Do {\bf not} include this section in the anonymized submission, only in the final paper. You can use the \texttt{ack} environment provided in the style file to autmoatically hide this section in the anonymized submission.
\end{ack}

\small
\bibliography{all}
\bibliographystyle{abbrv}
\normalsize

\clearpage

\appendix

\section{Supplemental Material}
\label{sec:supplemental}

\subsection{Details of the Numerical Reasoning Task}\label{sec:syntext_data}
For creating the textual synthetic data for the generative QA task of \S\ref{sec:syntext}, we used the data generation set-up of \cite{ggb2020injecting} (\S4.2 of \cite{ggb2020injecting}). Default templates and vocabulary were used to create passages containing 5 sentences. While instantiating the templates, the probability of sampling from one of the previously used values was set to 0.999 to promote inter-sentence dependencies. This gave us 629906/15K train/dev passage-question-answer triples. Among these, we only kept the samples where the answer was a number not appearing in the passage, and discarded the rest. This gave us 223067 training and 5146 evaluation instances. 

We only kept the decoder/generative head of \genbert{} (\S3 of \cite{ggb2020injecting}) and allowed the decoder to attend to all the encoder outputs in the cross-attention layers. As the weights of the encoder and decoder are tied, we used segment ids 0 for the encoder input sequence and 1 for the decoder inputs. 

\subsection{Data for Masked LM task}\label{sec:mlm_data}
The instances for the MLM task (\S\ref{sec:lm}) were formed separately using 5.2M pages from English Wikipedia (October 2017 dump) and the training set of PG19 dataset containing $\sim$29K books from Project Gutenberg \cite{raecompressive2019}. For each dataset, after appending a special symbol at the end of each document, the documents were arranged in a random order and concatenated into a single long text which was then tokenized into a list of tokens. Depending upon the input length $L$ of the experiment (512/1024/etc) this list was chunked into full length $L-2$ sequences which were then masked randomly following \cite{devlin2018bert} and enclosed within \texttt{[CLS]} and \texttt{[SEP]} tokens. For each dataset, the first 2.55B tokens (i.e. $510\times5$M) were used to form the respective training set, next 10.20M tokens ($510\times20$K) the dev set and the rest were discarded. 

\subsection{Finetuning on \hotpotqa{}}\label{sec:hotpot-details}

Given the question $Q$ and 10 arranged paragraphs $P_i$'s, each $P_i$ is extended by prefixing it with its title. Moreover, to handle \textit{yes}/\textit{no} questions, a special string \texttt{<a> yes no </a>} is also prefixed. The context $D$ is formed by simply concatenating the resulting paragraphs. Following \cite{devlin2018bert}, given a window/chunk $P$ from tokenized $D$, the corresponding instance is formed as \texttt{[CLS] Q [SEP] P [SEP]}.  

\textit{Supporting Facts Tagging Task} (\textsc{SF}): Besides the standard span extraction loss, we also include another task using the supporting facts supervision. Contextualized representations of the model are linearly projected to 2 scores (for 0/1) per token and normalized to obtain log-probabilities. For an input, loss is computed as negative log probability of the correct tag averaged over the positions. As supporting facts positions are fewer, log-probabilities are weighted according to the respective class (0/1) size.

\textit{Adversarial data generation:} After training the single-paragraph model of \cite{min2019compositional} on \hotpotqa{}, for each sample in the training and development sets, we retrieved top 50 introductory paragraphs from Wikipedia according their TF-IDF similarity with the question. The 50 paragraphs were then re-ranked using the "no-answer-logit" score predicted by the trained model and 8 adversarial distractors were chosen accordingly. When evaluated on the adversarial version of the development set the performance of the trained model reduced from $64.4 \rightarrow 57.8$ \fone{}. Re-training on the adversarial data increased the performance to $61.3$. In both cases, we trained for 10 epochs with batch size 36, maximum sequence length 300 and learning rate 5e-5 with linear warmup proportion 0.1.

\subsection{Hyperparameters}\label{sec:exp_setup}
For all our experiments, we used an older version of Hugging Face's Transformers library \cite{Wolf2019HuggingFacesTS}. For convenience, we denote the training hyperparameters using the following abbreviations, INS: number of training instances, BSZ: number of instances in a batch, ISZ: instance size, SQL: final input sequence length, LR: learning rate, WRM: linear LR warm-up proportion, EP: number of epochs, STP: number of optimizer steps, GAC: gradient accumulation steps, POSq: whether (y/n) $q$ part is included in positional embeddings defined in \S\ref{sec:model}. 

The hyperparameters for majority tagging are in Table~\ref{table:hyperparams-majority}, for \genbert{} finetuning in Table~\ref{table:hyperparams-syntext}, for MLM trainings in Table~\ref{table:hyperparams-mlm}, for \squad{} finetuning in Table~\ref{table:hyperparams-squad} and for \hotpotqa{} finetuning in Table~\ref{table:hyperparams-hotpot}.

\begin{table}[h!]
    \scriptsize
    \centering
    \begin{tabular}{|c|c|c|c|c|c|c|c|c|c|c|c|} \hline
    % \begin{tabular}{cccccccccccc} \hline
        section & model(s) & init & BSZ & ISZ & SQL & LR & WRM & EP & STP & GAC & POSq \\\hline
        \S\ref{sec:rand_init} & $(512\times 1, 0)$ & random & 60 & 512 & 512 & 5e-5 & 0.1 & 9 & 431K & 1 & n \\\hline
        \S\ref{sec:rand_init} & $(512\times 4, 64)$ & random & 60 & 512 & 2048 & 5e-5 & 0.1 & 9 & 431K & 1 & y \\\hline
        \S\ref{sec:rand_init} & $(1024\times 1, 0)$ & random & 60 & 512 & 1024 & 5e-5 & 0.1 & 9 & 431K & 1 & y \\\hline
        \S\ref{sec:bert_init} & \begin{tabular}[c]{@{}c@{}c@{}c@{}} $(512\times 4, 0)$\\ $(512\times 4, 64)$ \\ $(1024\times 2, 0)$ \\ Wikipedia , PG19 \end{tabular} & \bert{} & 16 & 512 & 2048 & 2e-5 & 0.001 & 5 & 350K & 1 & y \\\hline
        \S\ref{sec:bert_init} & $(8\times 64, 0)$ & $(512\times 4, 0)$ & 24 & 512 & 512 & 5e-5 & 0.01 & 5 & 270K & 1 & y \\\hline
        \S\ref{sec:bert_init} & $(8\times 64, 64)$ & $(512\times 4, 64)$ & 24 & 512 & 512 & 5e-5 & 0.01 & 5 & 270K & 1 & y \\\hline
        \S\ref{sec:compress_eval} & \begin{tabular}[c]{@{}c@{}} $(512\times 1, 64)$, $N_c=3$ \\ $(512\times 1, 64)$, $N_c=9$ \end{tabular} & $(512\times 4, 64)$ & 24 & 512 & 512 & 5e-5 & 0.01 & 5 & 440K & 1 & y \\\hline
    \end{tabular}
    \caption{Training hyperparameters for MLM training (\S\ref{sec:lm}). Common parameters: INS=5M, dropout-rate=0.1, optimizer=Bert-Adam, weight-decay=0.01, max-grad-norm=1.0, seed=42. If STP specified, training is terminated after STP-many optimizer steps.}
    \label{table:hyperparams-mlm}
\end{table}

\begin{table}[h!]
    \scriptsize
    \centering
    \begin{tabular}{|c|c|c|c|c|c|c|c|c|c|c|c|} \hline
    % \begin{tabular}{cccccccccccc} \hline
        model(s) & BSZ & ISZ & SQL & LR & WRM & EP & GAC  \\\hline
         all & 12 & 384 & 384 & 3e-5 & 0.1 & 3  &  1  \\\hline
    \end{tabular}
    \caption{Training hyperparameters for \squad{} v1 finetuning (\S\ref{sec:rc}). POSq for a model is same as during its pre-training. Common parameters: maximum query length=64, window-stride-length=128, dropout-rate=0.1, optimizer=Bert-Adam, weight-decay=0.01, max-grad-norm=1.0, seed=42.}
    \label{table:hyperparams-squad}
\end{table}

\begin{table}[h!]
    \scriptsize
    \centering
    \begin{tabular}{|c|c|c|c|c|c|c|c|c|c|c|c|} \hline
    % \begin{tabular}{cccccccccccc} \hline
        task & num. of layers & INS & BSZ & ISZ & SQL & LR & WRM & EP & GAC & POSq \\\hline
        L = 8192 , p = 1  & 2 & 200K & 4 & 8192 & 8192 & 2e-6 & 0.01 & 1 &  1 & y \\\hline
        L = 512 , p = 1, 3  & 2 & 200K & 80 & 512 & 512 & 2e-6 & 0.01 & 2 &  1 & n \\\hline
        L = 1792 , p = 4  & 12 & 300K & 8 & 1792 & 1792 & 2e-6 & 0.01 & 2 &  1 & y \\\hline
    \end{tabular}
    \caption{Training hyperparameters for majority tagging task (\S\ref{sec:majority}). Common parameters: init=random, dropout-rate=0.0, optimizer=Bert-Adam, weight-decay=0.01, max-grad-norm=1.0, seed=42.}
    \label{table:hyperparams-majority}
\end{table}

\begin{table}[h!]
    \scriptsize
    \centering
    \begin{tabular}{|c|c|c|c|c|c|c|c|c|c|c|c|} \hline
    % \begin{tabular}{cccccccccccc} \hline
        model(s) & BSZ & ISZ & SQL & LR & WRM & EP & GAC  \\\hline
         all & 68 & 140 & 140 & 3e-5 & 0.1 & 15  &  1  \\\hline
    \end{tabular}
    \caption{Training hyperparameters for \genbert{} finetuning (\S\ref{sec:syntext}). Common parameters: init=\bert{}, INS=223067, POSq=n, dropout-rate=0.1, optimizer=Bert-Adam, weight-decay=0.01, max-grad-norm=1.0, seed=42.}
    \label{table:hyperparams-syntext}
\end{table}

\begin{table}[h!]
    \scriptsize
    \centering
    \begin{tabular}{|c|c|c|c|c|c|c|c|c|c|c|c|} \hline
    % \begin{tabular}{cccccccccccc} \hline
        model(s) & dataset-setting & window-stride-length & BSZ & ISZ & SQL & LR & WRM & EP & GAC & POSq \\\hline
         \bert{} & \textsc{GS} & 128 & 32 & 512 & 512 & 2.5e-5 & 0.1 & 3  &  1  & n \\\hline
         \begin{tabular}[c]{@{}c@{}} $(512\times 4, 0)$ \\ $(512\times 4, 64)$ \end{tabular} & \textsc{GS} & 128 & 12 & 2048 & 2048 & 3.5e-5 & 0.1 & 6  &  1  & y \\\hline
         \begin{tabular}[c]{@{}c@{}} $(512\times 4, 0)$ \\ $(512\times 4, 64)$ \end{tabular} & $\textsc{GS}+\textsc{QR}$ & 256 & 8 & 2048 & 2048 & 3e-5 & 0.1 & 4  &  2  & y \\\hline
    \end{tabular}
    \caption{Training hyperparameters for finetuning on \hotpotqa{} variants (\S\ref{sec:rc}). Common parameters: maximum query length=64, dropout-rate=0.1, optimizer=Bert-Adam, weight-decay=0.01, max-grad-norm=1.0, seed=42.}
    \label{table:hyperparams-hotpot}
\end{table}

\end{document}